\documentclass{article}

\usepackage{aaai4nmr}
\usepackage{makeidx}
\usepackage[T1]{fontenc}
\usepackage{amsmath}
\usepackage{amsfonts}
\usepackage{latexsym}
\usepackage{graphicx}

\newcommand{\dft}[3]{\ensuremath{\frac{{#1}\,:\,{#2}}{#3}}}
\newcommand{\dr}[3]{\ensuremath{\frac{{#1}\,:\,{#2}}{#3}}}

\newcommand{\theory}[1]{\ensuremath{T\!h\!\left({#1}\right)}}
\newcommand{\default}{\dr{\alpha}{\beta_1,\dots,\beta_n}{\gamma}}

\newcommand{\CupEk}{\bigcup_{k = 0}^{\infty} E_k}

\begin{document}

\bibliographystyle{aaai}

\title{Description of GADEL}
\author{Igor St\'ephan, Fr\'ed\'eric Saubion and Pascal
Nicolas \\LERIA, Universit\'e d'Angers\\2 Bd Lavoisier\\F-49045 Angers Cedex 01\\{\small \tt \{Igor.Stephan,Frederic.Saubion,Pascal.Nicolas\}@univ-angers.fr}
}

\maketitle

\begin{abstract}
This article describes the first implementation of the GADEL system : a Genetic
Algorithm for Default Logic.
The goal of GADEL is to compute extensions in Reiter's default logic.
It accepts every kind of finite propositional default theories and
is based on evolutionary principles of Genetic Algorithms.
Its first experimental results on certain instances of the problem
show that this new approach of the problem can be successful.

\end{abstract}
\section{General Info}

The system works on SUN/Solaris or PC/Linux with Sicstus Prolog3.7.x and C.
It is written in Prolog and it generates a Sicstus library in C.
The current version is about 3000 lines of Prolog.

\section{Description of the System}
\subsection{Default Logic and Genetic Algorithms}
{\em Default Logic} has been introduced by Reiter~\cite{reiter80} in
order to formalize common sense reasoning from incomplete information,
and is now recognized as one of the most appropriate framework for
{\em non monotonic reasoning}.  In this formalism, knowledge is
represented by a default theory $(W,D)$ where $W$ is a set of first order
formulas representing the sure knowledge, and $D$ a set of
\emph{default rules} (or defaults).  A \emph{default}
$\delta=\default{}$ is an inference rule providing conclusions relying
upon given, as well as absent information meaning ``if the
\emph{prerequisite} $\alpha$ is proved, and if for all $i=1,\dots,n$
each \emph{justification} $\beta_i$ is individually consistent (in
other words if nothing proves its negation) then one concludes the
\emph{consequent} $\gamma$''.
From a default theory $(W,D)$ one tries to build some
extensions, that are maximal sets of plausible conclusions.
Reiter has given the following
pseudo iterative characterization of an extension $E$:
we define 
\begin{itemize}
\item $E_0 = W$
\item and for all  $k \geq 0$, \\
\begin{eqnarray*}
   E_{k+1} &=& \theory{E_k}\cup \{\gamma \mid \default \in D,   \\
     && \alpha \in E_k,\neg\beta_i \not\in E,  \forall i=1,\dots,n \}
  \end{eqnarray*}
\end{itemize}
then, $E$ is an extension of $(W, D)$ iff $E = \CupEk$.
The computation of an extension is known to be
$\Sigma_2^p-complete$~\cite{gottlob92}. 
Even if the system DeRes~\cite{chmamitr99} has very good performance 
on certain classes of default theories, there is no efficient system for 
general extension calculus. 
The aim of the present work is to describe the first implementation of the
GADEL system (\emph{Genetic Algorithms for DEfault Logic})
which applies {\em Genetic Algorithms} principles to propositional default reasoning~\cite{NMR00}.

Based on the principle of natural selection, Genetic Algorithms
have been quite successfully applied to combinatorial problems such as
scheduling or transportation problems. The key principle of this
approach states that, species evolve through adaptations to a changing
environment and that the gained knowledge is embedded in the structure
of the {\em population} and its members, encoded in their {\em chromosomes}. 
If individuals are considered as potential solutions to a given problem,
applying a genetic algorithm consists in generating better and better
individuals.
A genetic algorithm consists of the following components:
\begin{itemize}
\item
a representation of the potential solutions in a chromosome, in most cases, a string of bits representing its {\em genes}, 
\item
an initial population,
\item
an {\em evaluation function} which rates each potential solution
w.r.t. the given problem,
\item
genetic operators that define the composition of the offsprings : two different
operators will be considered : {\em crossover} allows to generate two new chromosomes
(the offsprings) by crossing two chromosomes of the current population (the
parents), {\em mutation} arbitrarily alters one or more genes of a selected chromosome,
\item
parameters : population size $p_{size}$ and probabilities of crossover $p_c$
and mutation $p_m$.
\end{itemize}
and an iteration process:
\begin{itemize}
\item
evaluate each chromosomes,
\item
order the population according to evaluation rates and select the bests chromosomes,
\item 
perform crossover and mutation on pairs of randomly selected chromosomes,
\item repeat this full process until a user-defined number of populations has
been explored.
\end{itemize}
The best chromosome of each population w.r.t. the evaluation function
represents the current best solution to the problem.

Before a detail description of the GADEL system, it is necessary to give some
arguments about the choice of the implementation language:
\begin{itemize}
\item In the previous section we have presented the most common version of GA
but in fact each part of the system can take various forms. To develop easily
a GA system the implementation language must be very flexible and modular.
\item A GA system is an iterative system. The implementation language must be efficient.
\item Default Logic is based on classical logic. To develop easily a GA system
about Default Logic, the implementation language must be logic and symbolic.
\end{itemize}
From above, the choice of the implementation language is obvious: the most
popular and efficient of the logic programming language, Prolog.
We assume in the rest of the paper a minimal knowledge of Prolog.
The article is organized as follows: section 2 presents the genetic
algorithms aspects of GADEL, section 3 provides the process of compilation of
a default theory to a prolog program, section 4 focuses on the evaluation
function of GADEL and section 5 describes our experiments w.r.t. other
existing systems.

\subsection{GADEL: a GA system}

\subsubsection{Representation and semantics.}
Our purpose is to construct an
extension of a given default theory $(W,D)$. 
For each default $\default$ we encode in the chromosome the prerequisite
$\alpha$ and all justifications $\beta_1,...,\beta_n$ conjointly. 
Given a set of defaults of size $n$ the chromosome will be of size $2n$.
A {\em candidate extension} $CE(G)$ associated to a chromosome $G$ is :
\[CE(G) = Th(W \cup \left\{\begin{array}{l}\gamma_i \mid
\dft{\alpha_i}{\beta_i^1,...,\beta_i^{k_i}}{\gamma_i} \in D\\
 and \; G|_{2i-1}=1 \;and\; G|_{2i}=0 \end{array}\right\})
\]

\subsubsection{Population tree.}
According to the principles of Genetic Algorithms, we now consider a
population of individuals representing candidate extensions.
Usually chromosomes are strings of bits and population are sets of
chromosomes.
We have chosen a new representation for the population: binary trees.
A population is defined inductively on the set of constructors $\Lambda$,
$zero$, $one$ and $zero-one$ of arity, respectively, 0, 1, 1, and 2.
The two advantages of this representation are its compactness and unicity 
of each chromosome.

For example, the population \(\{0101,1000,1010\}\) is represented by the tree
in Figure~1.

\begin{figure}
\small
\begin{center}
\setlength{\unitlength}{0.00083300in}%
\begingroup\makeatletter\ifx\SetFigFont\undefined
\def\x#1#2#3#4#5#6#7\relax{\def\x{#1#2#3#4#5#6}}%
\expandafter\x\fmtname xxxxxx\relax \def\y{splain}%
\ifx\x\y   
\gdef\SetFigFont#1#2#3{%
  \ifnum #1<17\tiny\else \ifnum #1<20\small\else
  \ifnum #1<24\normalsize\else \ifnum #1<29\large\else
  \ifnum #1<34\Large\else \ifnum #1<41\LARGE\else
     \huge\fi\fi\fi\fi\fi\fi
  \csname #3\endcsname}%
\else
\gdef\SetFigFont#1#2#3{\begingroup
  \count@#1\relax \ifnum 25<\count@\count@25\fi
  \def\x{\endgroup\@setsize\SetFigFont{#2pt}}%
  \expandafter\x
    \csname \romannumeral\the\count@ pt\expandafter\endcsname
    \csname @\romannumeral\the\count@ pt\endcsname
  \csname #3\endcsname}%
\fi
\fi\endgroup
\begin{picture}(2347,2589)(1077,-3694)
\thicklines
\put(1501,-3436){\line( 0, 1){300}}
\put(1501,-2836){\line( 0, 1){300}}
\put(1501,-2236){\line( 0, 1){300}}
\put(2401,-3436){\line( 0, 1){300}}
\put(3001,-3436){\line( 0, 1){300}}
\put(2401,-2836){\line( 1, 1){300}}
\put(2701,-2536){\line( 1,-1){300}}
\put(2701,-1936){\line( 0,-1){300}}
\put(1501,-1636){\line( 2, 1){600}}
\put(2101,-1336){\line( 2,-1){600}}
\put(1501,-1861){\makebox(0,0)[b]{\smash{\SetFigFont{12}{14.4}{rm}$one$}}}
\put(1501,-2461){\makebox(0,0)[b]{\smash{\SetFigFont{12}{14.4}{rm}$zero$}}}
\put(1501,-3061){\makebox(0,0)[b]{\smash{\SetFigFont{12}{14.4}{rm}$one$}}}
\put(1501,-3661){\makebox(0,0)[b]{\smash{\SetFigFont{12}{14.4}{rm}$\Lambda$}}}
\put(2701,-1861){\makebox(0,0)[b]{\smash{\SetFigFont{12}{14.4}{rm}$zero$}}}
\put(2701,-2461){\makebox(0,0)[b]{\smash{\SetFigFont{12}{14.4}{rm}$zero\_one$}}}
\put(2401,-3061){\makebox(0,0)[b]{\smash{\SetFigFont{12}{14.4}{rm}$zero$}}}
\put(3001,-3061){\makebox(0,0)[b]{\smash{\SetFigFont{12}{14.4}{rm}$zero$}}}
\put(2401,-3661){\makebox(0,0)[b]{\smash{\SetFigFont{12}{14.4}{rm}$\Lambda$}}}
\put(3001,-3661){\makebox(0,0)[b]{\smash{\SetFigFont{12}{14.4}{rm}$\Lambda$}}}
\put(2101,-1261){\makebox(0,0)[b]{\smash{\SetFigFont{12}{14.4}{rm}$zero\_one$}}}
\end{picture}
\end{center}
\caption{Population \(\{0101,1000,1010\}\)}
\end{figure}
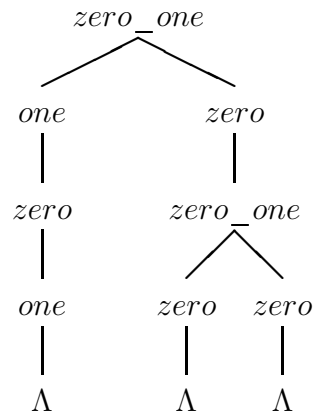

\subsection{Compilation of default theory}
A default theory is the given of a set of (propositional) formulas $W$ and a
set of defaults $D$. 
Prerequisite, conclusion and justifications of a default are all
(propositional) formulas.
So the GADEL system needs a classical theorem prover. 
It must be efficient because it is applied on each chromosome at each
new population.
The obvious choice is to compile all these sets of formulas into a set of
clauses. 
We have chosen the compilation to a disjunctive Prolog program.

\subsubsection{Small introduction to disjunctive logic programming.}
The theoretical basis of Prolog is the SLD-resolution for Horn
clauses\footnote{A Horn clause is a clause with at most one positive literal}.
It is not possible to directly insert disjunctive clauses in a Prolog
program\footnote{A clause is disjunctive if it contains at least two positive
literals}.
Disjunctive logic programming (resp. disjunctive Prolog) is an ``extension'' of Horn logic programming
(resp. Prolog) which allows disjunctions in the heads of definite\footnote{A
clause is definite if it contains one and only one positive literal} clauses
(resp. Prolog clauses).
A way to handle disjunctive clauses is the {\em case-analysis principle}:
a set of clauses $\{C\vee C', C_1,\ldots,C_m\}$ is unsatisfiable if and only if
the two sets of clauses $\{C,C_1,\ldots,C_m\}$ and $\{C', C_1,\ldots,C_m\}$ are
also unsatisfiable.
Disjunctive clauses $(h_1 \vee \ldots \vee h_p \vee \neg b_1 \vee \ldots \vee \neg
b_n)$ ($h_k$, $1\leq k\leq p$ and $b_k$, $1\leq k \leq n$ atoms) are then
written  $(h_1 \vee \ldots \vee h_p \leftarrow b_1 \wedge \ldots \wedge b_n)$.
We have chosen the SLOU-resolution approach~\cite{AIMSA98} (case-analysis and
SLD-reduction) as the theoretical basis of SLOU Prolog, our implementation of disjunctive Prolog.

The strategy of SLOU Prolog applies case-analysis by necessity: the
case-analysis is used only when a head $h_i$ of a disjunctive clause $(h_1 \vee
\ldots \vee h_p \leftarrow b_1 \wedge \ldots \wedge b_n)$ is useful in the proof.
This strategy needs that a negative clause\footnote{A clause is negative if
it contains only negative literals} $\neg b_1 \vee \ldots \vee \neg b_n$ is
written $false \leftarrow b_1 \wedge \ldots \wedge b_n$ and $false$ becomes
the goal. 
This choice assumes that almost all the clauses are definite (Prolog) clauses.

\subsubsection{From set of defaults to disjunctive Prolog program.}
Let us give a default $\frac{\alpha_i:
\beta^1_i,\ldots,\beta^{k_i}_i}{\gamma_i}$, the $i^{th}$ of the default theory
and a chromosome $G$.
If $G|_{2i-1}=1$ and $G|_{2i}=0$ then the default is supposed to be applied
and $\gamma$ must be added to the candidate extension.
Hence $G$ has to be a parameter of the theorem prover.
In order to calculate $(CE(G)\vdash \alpha_i)$ or $(\exists j(CE(G) \vdash \neg
\beta^j_i))$, $i$ and $(i,j)$ also have to be parameters of the theorem
prover.
So a propositional variable $h$ is compiled in a disjunctive Prolog atom
$h(I,G)$.
We can now describe the compilation of the three parts of a default rule.
\begin{itemize}
\item{Compilation of a conclusion:} 
In order to calculate $CE(G)$, we must add $\gamma_i$
to the candidate extension if $G|_{2i-1}=1$ and $G|_{2i}=0$.
We first normalize $\gamma_i$ in a set of disjunctive clauses  
$\{h_1\vee \ldots \vee h_p \leftarrow b_1 \wedge \ldots \wedge b_n\}$ 
and compile it in a disjunctive Prolog definition:
\begin{eqnarray*}
\lefteqn{\{h_1(I,G) ; \ldots ; h_p(I,G) :- }\\
&G|_{2I-1}=1,G|_{2I}=0,b_1(I,G),\ldots,b_n(I,G)\}
\end{eqnarray*}
\item{Compilation of a prerequisite:} 
The function $f$ of the GADEL evaluation function compares $G|_{2i-1}$
with  $CE(G) \vdash \alpha_i$ (see section~\ref{sec:eval}). 
To prove it with a disjunctive logic program we prove $CE(G), \neg
\alpha_i\vdash false$.
We first normalize $\neg \alpha_i$ in a set of clauses $\{h_1\vee \ldots \vee h_p \leftarrow b_1 \wedge \ldots \wedge b_n\}$
and compile it in a disjunctive Prolog definition:
\begin{eqnarray*}
\lefteqn{\{h_1(I,G) ; \ldots ; h_p(I,G) :-}\\
& I=i, b_1(I,G),\ldots,b_n(I,G)\}
\end{eqnarray*}

\item{Compilation of justifications:}
The function $f$ of the GADEL evaluation function compares $G|_{2i}$ with
$\exists j(CE(G)\vdash \neg \beta^j_i)$.
To prove it with a disjunctive logic programs we prove $(\exists
j(CE(G),\beta^j_i \vdash false))$.
We first normalize $\beta_i^j$ in a set of clauses $\{h_1\vee \ldots \vee h_p$$ \leftarrow b_1 \wedge \ldots \wedge b_n\}, \forall j$,
and  compile it in a disjunctive Prolog definition:
\begin{eqnarray*}
\lefteqn{\{h_1(I,G) ; \ldots ; h_p(I,G) :-}\\
& I=(i,j),b_1(I,G),\ldots,b_n(I,G)\}
\end{eqnarray*}

\item{Compilation of a formula $\omega\in W$:}
We first normalize $\omega$ in a set of clauses $\{h_1\vee \ldots \vee h_p$ $\leftarrow b_1 \wedge \ldots \wedge b_n\}$
and  compile it in a disjunctive Prolog definition:
\begin{eqnarray*}
\lefteqn{\{h_1(I,G) ; \ldots ; h_p(I,G) :-}\\
& b_1(I,G),\ldots,b_n(I,G)\}
\end{eqnarray*}
\end{itemize}

\subsubsection{From disjunctive Prolog to Prolog.}
During the execution of a disjunctive Prolog program, some clauses are
dynamic: case-analysis creates from a disjunctive clause $(h_1\vee \ldots \vee
h_p \leftarrow b_1 \wedge \ldots \wedge b_n)$  $p$ new clauses $(h_1
\leftarrow b_1 \wedge \ldots \wedge b_n)$ and $\{h_k\}_{1<k\leq p}$.
A disjunctive clause is called {\em usable} if it has not been splitted
in this set of clauses otherwise it is called {\em unusable}.
A new clause is usable if it is the result of a case-analysis.
To realize the case-analysis principle in Prolog, one needs to extend each
predicate with a {\em program continuation} $P$ (a difference list) that
handles those dynamic clauses.
A disjunctive Prolog clause \((h_1(I,G) ; \ldots ; h_p(I,G) :-\)
\(b_1(I,G),\ldots,b_n(I,G))\) is then compiled in a set of Prolog
clauses as follows:
\begin{eqnarray*}
\lefteqn{h_1(I,G,P) :-}\\
& & usable\_?(h_1\leftarrow b_1 \wedge \ldots \wedge b_n,P),\\
&&b_1(I,G,P),\ldots,b_n(I,G,P)\\
&;&\\
&& usable\_?(h_1\vee \ldots \vee h_p \leftarrow b_1 \wedge \ldots \wedge b_n,P),\\
&& unusable\_!(h_1\vee \ldots \vee h_p \leftarrow b_1 \wedge \ldots \wedge
b_n,P),\\
& \{&same\_assumptions(P,P_l),\\&&usable\_!(h_l ,P_l),
false(I,G,P_l),\}_{1<l\leq p}\\
&& usable\_!(h_1 \leftarrow b_1 \wedge \ldots \wedge b_n,P), \\
&&b_1(I,G,P),\ldots,b_n(I,G,P).\\
\lefteqn{\{h_k(I,G,P) :-}\\
& & usable\_?(h_k,P)\\
&;&\\
&& usable\_?(h_1\vee \ldots \vee h_p \leftarrow b_1 \wedge \ldots \wedge b_n,P),\\
&& unusable\_!(h_1\vee \ldots \vee h_p \leftarrow b_1 \wedge \ldots \wedge
b_n,P),\\
&\{&same\_assumptions(P,P_l),\\&&usable\_!(h_l,P_l), false(I,G,P_l),\}_{1\leq l\leq p, l\neq k}\\
&& usable\_!(h_k \leftarrow b_1 \wedge \ldots \wedge b_n,P).\}_{1<k\leq p}
\end{eqnarray*}

\subsection{Evaluation function of GADEL}
\label{sec:eval}
The evaluation function is the heart of the GADEL system.
It rates each chromosome given a default theory compiled in a disjunctive
prolog program.

\subsubsection{Evaluation of pair of genes.}
For a default $\delta_i = \dft{\alpha_i}{\beta_i^1,...,\beta^{k_i}_i}{\gamma_i}$,
an intermediate evaluation function $f$ is defined in Table~\ref{eval}. Given the two positions $G|_{2i-1}$ and $G|_{2i}$ in the
chromosome associated to the default $\delta_i$, the first point is to
determine w.r.t. these values if this default is supposed to be
involved in the construction of the candidate extension and then to check if
this application is relevant. 

\begin{table}[htbp]
\small\[
\begin{array}{|c|c|c|c|c|}
\hline
G|_{2i-1} & G|_{2i} & CE(G) \vdash \alpha_i & \exists j, CE(G) \vdash \neg
\beta_i^j & \hbox{penality} \\
\hline
1 & 0 & true & false & 0 \\
1 & 0 & true & true & p_2 \\
1 & 0 & false & true & p_3 \\
1 & 0 & false & false & p_4 \\
1 & 1 & true & false & p_5 \\
1 & 1 & true & true & 0 \\
1 & 1 & false & true & 0 \\
1 & 1 & false & false & 0 \\
0 & 1 & true & false & p_9 \\
0 & 1 & true & true & 0 \\
0 & 1 & false & true & 0 \\
0 & 1 & false & false & 0 \\
0 & 0 & true & false & p_{13} \\
0 & 0 & true & true & 0 \\
0 & 0 & false & true & 0 \\
0 & 0 & false & false & 0 \\
\hline
\end{array}
\]
\caption{Evaluation}
\label{eval}
\end{table}
We only illustrate $f$ on the case $G|_{2i-1}=1$ and  $G|_{2i}=0$ of the
Table~\ref{eval} (with penalty $p_2$ and a default with only one justification):
\begin{eqnarray*}
\lefteqn{f(I, (G,Eval\_G_{>I}),(G,Eval\_G_{\geq I})):-}\\
&&G|_{2I-1}=1, G|_{2I}=0,\\
&&false(I,G,\_P), \;\;\;\;\;\;\;\;\;\%\% (CE(G)\vdash \alpha_i) = true\\
&&false((I,1),G,\_P), \;\;\;\%\% (CE(G) \vdash \neg \beta_i)=true\\
&&Eval\_G_{\geq I} \;is\; Eval\_G_{>I} + p_2.
\end{eqnarray*}

\subsubsection{Evaluation of chromosome and population.}
The evaluation of a chromosome is the total sum of the evaluations for each
pair of genes.
Our evaluation function is calculated directly over the population tree by a
depth-first traversal.
The result is a set of pairs of a chromosome and its evaluation.
During the traversal, the construction of the current evaluated chromosome is
prefixed and the evaluation of the genes is postfixed.

\begin{eqnarray*}
\lefteqn{evaluation(Current\_Pop,Evaluated\_Pop):-}\\
&&eval\_alpha(0,\emptyset,Current\_Pop,Evaluated\_Pop).\\
\\
\lefteqn{eval\_alpha(I,G,one(Subtree),Evaluations) :-}\\
&&G|_{2I-1}=1,\\
&& eval\_beta(I,G,Subtree,Evaluations).\\
\\
\lefteqn{eval\_beta(I,G,zero(Subtree),Evaluations') :-}\\
&&G|_{2I}=0,\\
&&I' \;is\; I + 1,\\
&&eval\_alpha(I',G,Subtree,Evaluations),\\
&&map(f,I, Evaluations,Evaluations').
\end{eqnarray*}
\section{Applying the System}
\subsection{Methodology}
Methodology for using GADEL is the same as using default logic as a framework for knowldege representation.
\subsection{Specifics}
The semantics of our system is the Reiter's propositional default logic.
\subsection{Users and Useability}
The GADEL system takes default theory as Prolog facts in an input file.
Classical formulas of default theory are arbitrary formulas with
conjunctions, disjunctions and negations (noted resp. $\&\&$, $||$ and $!$).
A default is a triplet composed of a prerequisite, a list of justifications
and a conclusion.
GADEL is a framework for non monotonic reasoning systems. To extend GADEL to
an other system, one needs to redefine the evaluation function.
\section{Evaluating the System}
\subsection{Benchmarks}
We define two kinds of benchmarks: a taxonomic default theory ``\emph{people}'' described in
Table 2 and the well known Hamiltonian cycle problem in Table 5 as it has been described
and encoded in~\cite{chmamitr99}.
\subsection{Comparison}
\label{subsec:expresu}
DeRes and GADEL are compared on our two kinds of benchmarks.
CPU times given are in seconds on a SUN E3000 $(2 \times 250 Mhz)$.
The parameters of the genetic algorithm are $p_c=0.8$ and $p_m=0.1$.

\subsubsection{GADEL:1/DeRes:0.}

\begin{table}[htbp]

\begin{center}
{\small
\[\begin{array}{l}
W_{people} = \\
  \left\{  
    \begin{array}{ll}
      \neg boy \vee \neg girl, & \neg boy \vee kid,\\
         \neg girl \vee kid,  &\neg human \vee male \vee female, \\
         \neg kid \vee human,& \neg student \vee human,\\
      \neg adult \vee human,&\neg adult \vee \neg kid,\\
      \neg adult \vee \neg male \vee man,& \neg adult \vee \neg female \vee
  woman,\\
      \neg academic \vee adult,&\neg academic \vee diploma,\\
      \neg doctor \vee academic,& \neg priest \vee academic,\\
      \neg prof \vee academic,& \neg bishop \vee priest,\\
      \neg cardinal \vee bishop,& \neg redsuit \vee suit,\\
      \neg whitesuit \vee suit,& \neg blacksuit \vee suit,\\
      \neg redsuit \vee \neg whitesuit,& \neg whitesuit \vee \neg blacksuit,\\
       \neg redsuit \vee \neg blacksuit
    \end{array}
  \right\} \\
  \cup \{boy\}  or \cup \{girl\}  or \cup \{man\}  or \cup \{woman\} 
\cup \{man, student\}   \\or \cup \{woman, student\} \\
\\
D_{people} =\\
\left\{  
  \begin{array}{ll}
        \dr{human}{name}{name}&
        \dr{kid}{toys}{toys }\\
        \dr{student}{adult}{adult}&
        \dr{student}{\neg employed}{\neg employed}\\
        \dr{student}{\neg married}{\neg married}&
        \dr{student}{sports}{sports}\\
        \dr{adult}{\neg student}{employed}&
        \dr{adult}{\neg student, \neg priest}{married}\\
        \dr{adult}{car}{car}&
        \dr{adult}{\neg academic}{\neg toys}\\
        \dr{man}{\neg prof}{beer}&
        \dr{man}{\neg vegetarian}{steak}\\
        \dr{man}{coffee}{coffee}&
        \dr{man \vee woman}{wine}{wine}\\
        \dr{woman}{tea}{tea}&
        \dr{academic}{\neg prof}{\neg employed}\\
        \dr{academic}{\neg priest}{toys}&
        \dr{academic}{books}{books}\\
        \dr{academic}{glasses}{glasses}&
        \dr{academic}{\neg priest}{late}\\
        \dr{doctor}{medicine}{medicine} &
        \dr{doctor}{whitesuit}{whitesuit}\\
        \dr{prof}{employed}{employed}&
        \dr{prof}{grey}{grey}\\
        \dr{prof}{tie}{tie}&
        \dr{prof}{water}{water}\\
        \dr{prof}{conservative}{conservative}& 
        \dr{priest}{male}{male}\\
        \dr{priest}{conservative}{conservative}&
        \dr{priest}{\neg cardinal}{blacksuit}\\
        \dr{cardinal}{redsuit}{redsuit}&
        \dr{car}{mobile}{mobile}\\
        \dr{tie}{suit}{suit}&
        \dr{wine \wedge steak \wedge coffee}{\neg sports}{heartdisease}\\
        \dr{sports}{man}{football \vee rugby \vee tennis}&
        \dr{sports}{woman}{swim \vee jogging \vee tennis}\\
        \dr{toys \wedge (football \vee rugby)}{ball}{ball}&
        \dr{toys}{boy}{weapon}\\
        \dr{toys}{girl}{doll}
  \end{array}
\right\}  
\end{array}
\]
}

\caption{The taxonomic problem : ``people''}
\end{center}
\end{table}

\begin{table}[htbp]
\small
\begin{center}
\begin{tabular}{|l|r|r|r|r|}
    \hline
    & \multicolumn{3}{c|}{GADEL} & DeRes \\
    \hline
    Problem & $p_{size}$& Number of & CPU & CPU\\
    && populations &time &time \\
    \hline
    $boy$ &325& 3.3 & 15.4 & >7200\\
    $girl$&325& 3.4 & 15.6& >7200\\
    $man$&325& 3.5 & 25.3& >7200\\
    $woman$ &325& 3.0 & 14.6& >7200\\
    $man \wedge student$&325& 186.7 & 467.5 & >7200 \\
    $woman \wedge student$&325& 271.6 & 704.4 & >7200\\
    \hline
  \end{tabular}
\caption{GADEL:1/DeRes:0}
\end{center}
\end{table}

\begin{table}[htbp]
\small
\begin{center}
\begin{tabular}{|l|l|l|l|l|}
\hline
$N$&$p_{size}=$&CPU time& Number of&CPU time\\
&$\frac{N(N+1)}{2}$&&populations&\\
\hline
15&120 &2.6&6.6&17.2\\
16&136 &2.8&6.9&19.7\\
17&153 & 3.4&4.9&17.0\\
18&171 & 3.8&5.0&19.0\\
19&190&4.2&5.0&21.2\\
20&210 &4.7&4.5&21.3\\
21&231 & 4.9&4.8&23.8\\
22&253 & 5.6&3.9&22.1\\
23&276 &5.8&4.7&27.1\\
24&300 & 6.0&5.0&30.3\\
25&325 &7.2&3.5&25.3\\
\hline
\end{tabular}
\caption{$W_{people}\cup\{man\}$}
\end{center}

\begin{center}
\small
\begin{tabular}{|l|r|r|r|r|}
    \hline
    & \multicolumn{3}{c|}{GADEL} & DeRes \\
    \hline
    Problem &$p_{size}$ &Number & CPU & CPU \\
    && of pop &time &time \\
    \hline
    $ham.b\_3,2,0,0,1,0,0\_$ &465& 1.8 &5.6 & 0.5\\
    $ham.b\_4,2,0,0,1,0,0\_$ &465&- &>7200 & 19.4\\
    $ham.b\_5,2,0,0,1,0,0\_$ &465&- &>7200 & 566.4\\
    $ham.b\_6,2,0,0,1,0,0\_$ &465&- &>7200 &  >7200\\
    \hline
  \end{tabular}
\caption{GADEL:1/DeRes:1}
\end{center}

\end{table}

Table 3 gives results about the people default theory.
Each line corresponds to the common part of $W_{people}$ augmented
with one of the specified formula of the first column.
The second column  gives $p_{size}$ the initial number of chromosomes in the
popu\-lation, the third one is the average number of populations needed to
find an extension. The last two columns give CPU times.
\cite{chmamitr99} describes the very good performances of DeRes on some kind
of default theories: the stratified ones.
But it is also noticed that for a non stratified default theory the
performance of DeRes are not enough to deal with a non very few number of defaults.  
Results given in this table shows that DeRes has a lot of difficulties
with our taxonomic people example (even if the local prover is used).  
Conversely the number of populations are quite small for
GADEL (even if the time is not so good: all the implementation
is written in Prolog). 

Table 4 gives results about $W_{people}\cup\{man\}$ with different sizes of
populations (200 tests for each size of population). The second line gives
$p_{size}$ the initial number of chromosomes in the population. 
The third one gives the time spent for one complete computation of a new population.
The fourth one gives the average number of populations needed to find an
extension. 
The last one gives the average time to find an extension.
These results demonstrate that the size of the population must be balanced by
the time spent for one complete computation of a new population.
The increase of the population size does not necessarily increase the
efficiency of the genetic algorithm.
Finally, Figure~2 presents for $W_{people}\cup\{man\}$ with $N=17$ the number
of tests w.r.t. the number of populations needed to obtain an extension.
This figure suggests to stop computation after 6 populations and to
restart with a brand new one since $80\%$ of tests end after 6 populations at most.

\begin{figure}[htbp]
\begin{center}
\rotatebox{270}{\includegraphics[scale=0.25]{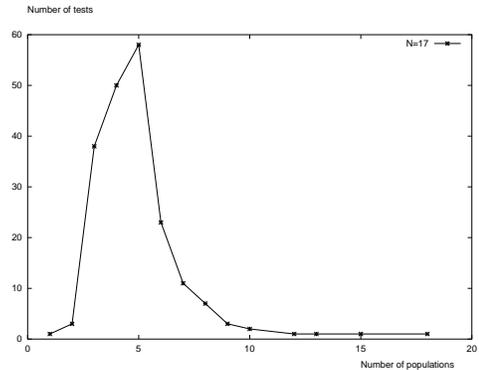}}
\end{center}
\caption{$W_{people}\cup\{man\}$ with $N=17$ ($p_{size}=153$)}
\end{figure}

\subsubsection{GADEL:1/Deres:1.}
GADEL has poor performances on Hamiltonian problems. 
We think that it is because we do not take
into account the groundedness~\cite{schwind90} into our evaluation function. 
As a matter of fact, in the Hamiltonian problem, a solution is exactly one
``chain''\footnote{We say that $\delta$ is chained to $\delta'$ if
  the prerequisite of $\delta'$ is deducible from $W$ and the consequent of $\delta$.} of defaults, but, there
is a lot of potential solutions (whose evaluation is null) based on
two, or more, chains of defaults. The only criterion to discard these
candidate extensions is the groundedness property that
they do not satisfy.  Conversely, in people example, a solution is a
set of non conflicting defaults, but at most four defaults are chained
together, and so the groundedness property is less important to reach
a solution. We are now testing some new evaluation functions in order to take
into account this criterion.

\subsubsection{Other systems.}
We have also in mind that in the area of logic programming and non
monotonic reasoning there exist others systems
(Smodels~\cite{niemela97}, DLV~\cite{eitleo98}) able to compute stable
models of extended logic program. Since this task is equivalent to
compute an extension of a default theory it seems interesting to
compare GADEL to these systems. But, even if DLV has the advantage
to accept formulas with variables which are instantiated before computation,
this system does not accept theories like our people example. On its part,
Smodels does not deal with this default theory because it can not be
represented by a normal logic program without disjunction. Because we have the
objective to deal with every kind of propositional formulas, GADEL spends a
lot of time in theorem proving and it seems not realistic to compare it with
those two systems. 
But it will be very inter-resting to work on GADEL's architecture in order to
improve its performances on particular subclasses of default theories. 

\subsection{Problem Size}
The system is a prototype which can handle non stratified theories with about
one hundred defaults.

\section{Conclusion}
\label{sec:conclu}
In this paper, we have described the first implementation of our
system GADEL whose goal is to compute extensions of every kind of
finite propositional Reiter's default theories.  Our new approach,
using principles of genetic algorithms, seems to be relevant as it is
illustrated by our experimental results.  But this present work is a
first approach and we have in mind many improvements as : more
accurate definition of the evaluation function, using reparation
techniques, local search heuristics.

\bibliography{/home/info/helios/pn/Tex/biblio,/home/info/helios/saubion/bibliographies/biblio,/home/info/helios/stephan/BIBLIOGRAPHIE/moi,/home/info/helios/stephan/BIBLIOGRAPHIE/slou}

\end{document}